\documentclass{acm_proc_article-sp}

\usepackage{times}
\usepackage{helvet}
\usepackage{courier}
\usepackage{times}
\usepackage{amssymb}
\usepackage{amsmath}
\usepackage{amssymb}
\usepackage{graphicx}
\usepackage{epsfig}
\usepackage{tikz}
\usepackage{balance}
\usetikzlibrary{shapes,arrows}
\usetikzlibrary{matrix}

\newtheorem{definition}{Definition}
\usepackage[lined,algoruled,linesnumbered]{algorithm2e}
\usepackage{subfigure}
\usepackage{amssymb}
\usepackage{amsmath}
\usepackage{multirow}
\begin{document}
%
\title{RDF2Rules: Learning Rules from RDF Knowledge Bases by  Mining Frequent Predicate Cycles}

\numberofauthors{2}
\author{
%
%
\alignauthor
Zhichun Wang\\
       \affaddr{College of Information Science and Technology, Beijing Normal University}\\
       \affaddr{Beijing, China}\\
       \email{zcwang@bnu.edu.cn}
\alignauthor
Juanzi Li\\
       \affaddr{Department of Computer Science and Technology, Tsinghua University}\\
       \affaddr{Beijing, China}\\
       \email{lijuanzi@tsinghua.edu.cn}
}

\maketitle
\begin{abstract}
Recently, several large-scale RDF knowledge bases have been built and applied in many knowledge-based applications. To further increase the number of facts in RDF knowledge bases, logic rules can be used to predict new facts based on the existing ones. Therefore, how to automatically learn reliable rules from large-scale knowledge bases becomes increasingly important. In this paper, we propose a novel rule learning approach named RDF2Rules for RDF knowledge bases. RDF2Rules first mines frequent predicate cycles (FPCs), a kind of interesting frequent patterns in knowledge bases, and then generates rules from the mined FPCs. Because each FPC can produce multiple rules, and effective pruning strategy is used in the process of mining FPCs, RDF2Rules works very efficiently. Another advantage of RDF2Rules is that it uses the entity type information when generates and evaluates rules, which makes the learned rules more accurate. Experiments show that our approach outperforms the compared approach in terms of both efficiency and accuracy.

\end{abstract}

\section{Introduction}
\label{sec:intro}

\noindent Recently, a growing number of large-scale Knowledge Bases (KBs) have been created and published by using the Resource Description Framework (RDF)\footnote{http://www.w3.org/RDF/}, such as DBpedia~\cite{bizer2009dbpedia}, YAGO~\cite{suchanek2008yago}, and Freebase~\cite{bollacker2008freebase} etc. These KBs contain not only huge number of entities but also rich entity relations, which makes them successfully used in many applications such as Question Answering~\cite{Unger:2012}, Semantic Relatedness Computation~\cite{leal2012computing} and Entity Linking\cite{shen2012linden}.

The coverage of entities and the amount of facts are two important factors that determine the quality of RDF KBs. In order to enrich the knowledge in an RDF KB, information extraction techniques are usually used to extract more entities and their relations from plain text or semi-structured text. For example, DBpedia regularly extracts facts from Wikipedia's infoboxes to update its contents. Yet another promising way to expand a KB is to infer new facts from the existing ones by using inference rules. For example, by using the following rule we can predict that entity $B$ is the child of entity $C$ if we have already known that $A$ is the parent of $B$ and $A$ is the spouse of $C$.
\[hasChild(A,B)\land hasSpouse(A,C)\Rightarrow hasChild(C,B)\]
Although the coverage of entities in a KB cannot be expanded in this way, inferring new facts by rules is more efficient and accurate than information extraction, especially when the KB has already accumulated substantial facts about entities. It is reported that the new version of YAGO used logic rules to deduce new facts from the existing ones~\cite{Hoffart201328}.

One challenging problem of inferring new facts is how to define all the possible inference rules for a KB. Since manually setting all the rules is not possible, how to automatically learning inference rules from the existing facts in KBs becomes an interesting and important problem.  Learning rules from KBs has been studied for years in the domain of Inductive Logic Programming (ILP)~\cite{Muggleton94inductivelogic}, but ILP approaches usually need negative facts of target relations and typically do not scale well on large-scale KBs. Recently, Gal{\'a}rraga et al. have proposed a system AMIE for learning rules from RDF data~\cite{galarraga2013amie}. AMIE starts with the most general rules having only heads, and gradually extends rules by using four operators. Each time an operator is executed, a projection query is submitted to the KB to select entities and relations for that operator. Most recently, AMIE has been Extended to AMIE+ by a series of improvements to make it more efficient~\cite{amieplus}. Compared with several state-of-the-art ILP systems, AMIE+ runs much more efficiently and can generate more rules with high quality.

In spite of good performance of AMIE+, there are still several challenging problems that need to be studied. First, how to use entity type information when learning rules has not been well studied. In~\cite{amieplus}, the authors do discuss adding types in rules, but how to automatically learn rules with types is not detailedly explained in the paper; and according to our experiments, the released AMIE+ tool can not learn rules with type information. Second, in the manner of learning one rule at a time, AMIE+ still needs a very long running time when the RDF KB is really large and long rules are to be learned. Third, the PCA confidence used by AMIE+ sometimes over-estimates unknown facts as true ones; rules with high PCA confidence may predict lots of incorrect facts.

In this paper, we propose a new rule learning approach named RDF2Rules. RDF2Rules works in a very different way from AMIE+ when learning inference rules. Instead of learning one rule at a time, RDF2Rules first mines a kind of interesting frequent patterns in KBs, which are called Frequent Predicate Cycles (FPCs); then multiple rules are generated from each mined FPC. With properly designed pruning strategy, RDF2Rules can running faster than AMIE+ does. In addition, RDF2Rules can use entity type information when generating and evaluating learned rules, which results in rules having more accurate predictions.%
 Specifically, our work has the following contributions:
\begin{itemize}
\item We introduce the concept of \textit{Frequent Predicate Cycle (FPC)} in RDF KBs, and manage to show that FPCs have corresponding relations with inference rules. An efficient algorithm for mining FPCs from RDF data is proposed; effective prune strategy is proposed to ensure the mining efficiency, and our FPC mining algorithm supports parallel execution on multi-core machines.
\item We propose a method for generating rules from the mined FPCs. Entity type information is utilized when our approach generates rules, and rules with entity type constraints are produced automatically.
\item To precisely evaluate the reliability of rules, we design a new confidence measure to evaluate rules under the open world assumption. Our new confidence measure also takes the entity type information into account, and can evaluate rules more accurately.
\item We evaluate our approach on YAGO2 and DBpedia. The experimental results show that our approach runs more efficiently and gets more reliable rules than the compared approach.
\end{itemize}

The rest of this paper is organized as follows, Section 2 first introduces some preliminary knowledge and then defines the concept of FPC; Section 3 presents the FPC mining algorithm; Section 4 presents the rule generation and evaluation methods; Section 5 describes the RDF KB indexing methods in our approach; Section 6 presents the evaluation results; Section 7 discusses some related work and finally Section 8 concludes this work.

\section{Frequent Predicate Cycles}
\label{sec:fpc}

In this section, we first introduce some basics of RDF KBs; and then define the concept of  \textit{Frequent Predicate Cycles} (FPCs), and discuss the relation between FPCs and logic rules.
\subsection{RDF and RDF KB}
\begin{figure}
    \begin{center}
        \begin{tikzpicture}
        \tikzstyle{entity}=[ellipse,thick,draw=blue!75,fill=blue!20,text height=6pt,font=\small]
        \tikzstyle{literal}=[rectangle,draw=red!75,fill=red!20,minimum height=8mm,font=\small]
        \tikzstyle{pre}=[<-,shorten <=1pt,>=stealth',semithick]
        \tikzstyle{post}=[->,shorten <=1pt,>=stealth',semithick]
          \node (Michelle) at ( -1,2) [entity] {dbr:Michelle\_Obama};
          \node (Biden) at ( 1,-2) [entity] {dbr:Joe\_Biden};
          \node (nameOfObama) at ( -3.5,-1.5) [literal] {"Barack Obama"@en};
          \node (Obama) at ( 0,0) [entity] {dbr:Barack\_Obama}
              edge [post]  node [swap,text height=6pt,font=\small] {dbo:spouse}  (Michelle)
              edge [post]  node [swap,text height=6pt,font=\small] {dbo:vicePresident}  (Biden)
              edge [post]  node [swap,text height=6pt,font=\small] {dbo:name}  (nameOfObama);
        \end{tikzpicture}
    \end{center}
    \caption{\small An example RDF graph for the resource "Barack\_obama"}
    \label{fig:rdf_graph_sample}
\end{figure}
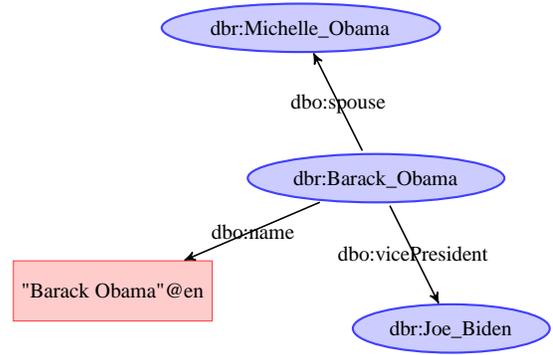

The Resource Description Framework (RDF) is a framework for the conceptual description or modeling of information in Web resources. RDF expresses information by making statements about resources in the form of
\[\langle subject\rangle \langle predicate\rangle \langle object \rangle.\]
The $subject$ and the $object$ represent two resources, the $predicate$ represents the relationship (directional) between the $subject$ and the $object$.  RDF statements are called triples because they consist of three elements. RDF is a graph-based data model; a set of RDF triples constitutes an RDF graph, where nodes represent resources and directed vertices represent predicates. There can be three kinds of nodes (resources) in an RDF graph: IRIs, literals, and blank nodes. An IRI is a global identifier for a resource, such as people, organization and place; literals are basic values including strings, dates and numbers, etc.; blank nodes in RDF represent recourses without global identifiers. Predicates in RDF are also represented by IRIs, since they can be considered as resources specifying binary relations. Figure \ref{fig:rdf_graph_sample} shows an example of small RDF graph built from three triples about Barack Obama in DBpedia; \textit{dbr} and \textit{dbo} stand for the IRI prefixes \textit{http://dbpedia.org/resource/} and \textit{http://dbpedia.org/ontology/}, respectively.

An RDF KB is a well-defined RDF dataset that consists of RDF statements (triples). The statements in an RDF KB are usually divided into two groups: T-box statements that define a set of domain specific concepts and predicates, and A-box statements that describe facts about instances of the concepts. The A-box triples excluding triples with literals are used by our approach to learn inference rules. Unlike AMIE, our approach also takes triples having \textit{rdf:type} predicate as input. \textit{rdf:type} is a special predicate that is used to state that a resource is an instance of a concept. The entity type information specified by \textit{rdf:type} predicate is very useful and important to rule learning from RDF KBs, which is verified by our experiments.

\subsection{Definitions}

RDF is a graph-based data model, so we represent an RDF KB as a graph $G=(E, P, T)$, where $E$ is the set of vertexes representing all the vertexes (entities), $P$ is the set of predicates, and $T \subseteq E\times P \times E$ are directed and labeled edges between vertexes (entities). Based on this graph representation of KB, we define the concept of \textit{Path} and \textit{Predicate Path} as follows.

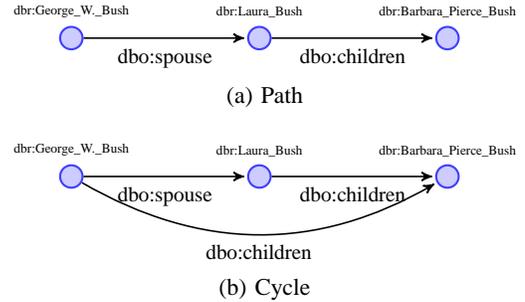
\begin{figure}
\label{fig:path}
\centering

\subfigure[Path]{
\begin{minipage}[b]{0.5\textwidth}
\label{fig:path}
\centering
\begin{tikzpicture}
\tikzstyle{entity}=[circle,thick,draw=blue!75,fill=blue!20]
\tikzstyle{pre}=[<-,shorten <=1pt,>=stealth',semithick]
\tikzstyle{post}=[->,shorten <=1pt,>=stealth',semithick]
\node (bush) at (-2.5,0) [entity] {};
\node (laura) at (0, 0) [entity] {}
edge [pre]  node [below,text height=6pt,font=\small] {dbo:spouse}  (bush);
\node (barbara) at (2.5,0) [entity] {}
edge [pre]  node [below,text height=6pt,font=\small] {dbo:children}  (laura);

\node [above,font=\tiny] at (bush.north) {dbr:George\_W.\_Bush};
\node [above,font=\tiny] at (laura.north) {dbr:Laura\_Bush};
\node [above,font=\tiny] at (barbara.north) {dbr:Barbara\_Pierce\_Bush};
\end{tikzpicture}
\end{minipage}
}

\subfigure[Cycle]{
\begin{minipage}[b]{0.5\textwidth}
\label{fig:cycle}
\centering
\begin{tikzpicture}
\tikzstyle{entity}=[circle,thick,draw=blue!75,fill=blue!20]
\tikzstyle{pre}=[<-,shorten <=1pt,>=stealth',semithick]
\tikzstyle{post}=[->,shorten <=1pt,>=stealth',semithick]
\node (bush) at (-2.5,0) [entity] {};
\node (laura) at (0, 0) [entity] {}
edge [pre]  node [below,text height=6pt,font=\small] {dbo:spouse}  (bush);
\node (barbara) at (2.5,0) [entity] {}
edge [pre]  node [below,text height=6pt,font=\small] {dbo:children}  (laura)
edge [pre, bend left]  node [below,text height=6pt,font=\small] {dbo:children}  (bush);

\node [above,font=\tiny] at (bush.north) {dbr:George\_W.\_Bush};
\node [above,font=\tiny] at (laura.north) {dbr:Laura\_Bush};
\node [above,font=\tiny] at (barbara.north) {dbr:Barbara\_Pierce\_Bush};
\end{tikzpicture}
\end{minipage}
}

\caption{Examples of path and cycle in DBpedia}
\end{figure}

\begin{definition}[Path]
A path in an RDF KB $G=(E, P, T)$ is a sequence of consecutive entities and predicates  $(v_1, p_1^{d_1}, v_2, p_2^{d_2},$ $...,p_{k-1}^{d_{k-1}}, v_k)$, where $v_i\in E$, $p_i\in P$; $d_i\in\{1,-1\}$ denotes the direction of predicate $p_i$, if $d_i=1$ then $\langle v_{i}, p_i, v_{i+1} \rangle \in T$; otherwise, $\langle v_{i+1}, p_i, v_{i} \rangle \in T$. The length of a path is the number of predicates in it.
\end{definition}

Paths in an RDF KB show how entities are linked by various relations. Figure~\ref{fig:path} shows an example of path in DBpedia, it starts from Grorge W. Bush via Lara Bush and ends at Barbara Pierce Bush. As shown in some studies, the starting and ending entities of a path sometimes also have some interesting relations. In the example of Figure~\ref{fig:path}, Barbara Pierce Bush actually is the children of Grorge W. Bush according to the facts in DBpedia. If we add this relation to the original path, we get a special kind of path as shown in Figure~\ref{fig:cycle}, the \textit{cycle}.

\begin{definition}[Cycle]
A cycle in an RDF graph is a special path that starts and ends at the same node.
\end{definition}

Cycles in RDF graphs show very interesting connection patterns among entities. The connection pattern shown in Figure~\ref{fig:cycle} reflects how Bush's family members are connected together by different relations. This pattern usually also holds for another family. In order to represent the interesting connection patterns in RDF graphs, we introduct the concept of \textit{Predicate Path} and \textit{Predicate Cycle}.

\begin{definition}[Predicate Path]
A predicate path is a sequence of entity variables and predicates $(x_1, p_1^{d_1}, x_2,...,p_k^{d_k}, x_{k+1})$, where $d_i\in\{1,-1\}$ denotes the direction of predicate edge $p_i$.
\end{definition}

\begin{definition}[Predicate Cycle]
A predicate cycle is a special predicate path that starts and ends at the same entity variable.
\end{definition}

\begin{figure}
\label{fig:pppc}
\centering
\subfigure[Predicate path]{
\begin{minipage}[b]{0.5\textwidth}
\label{fig:pred-path}
\centering
\begin{tikzpicture}
\tikzstyle{entity}=[circle,thick,draw=black!75,fill=white!75,inner sep=1.5pt,font=\small]
\tikzstyle{pre}=[<-,shorten <=1pt,>=stealth',semithick]
\tikzstyle{post}=[->,shorten <=1pt,>=stealth',semithick]
\node (bush) at (-2.5,0) [entity] {};
\node (laura) at (0, 0) [entity] {}
edge [pre]  node [below,text height=6pt,font=\small] {dbo:spouse}  (bush);
\node (barbara) at (2.5,0) [entity] {}
edge [pre]  node [below,text height=6pt,font=\small] {dbo:children}  (laura);

\node [above,font=\small] at (bush.north) {$x_1$};
\node [above,font=\small] at (laura.north) {$x_2$};
\node [above,font=\small] at (barbara.north) {$x_3$};
\end{tikzpicture}
\end{minipage}
}

\subfigure[Predicate cycle]{
\begin{minipage}[b]{0.5\textwidth}
\label{fig:pred-cycle}
\centering
\begin{tikzpicture}
\tikzstyle{entity}=[circle,thick,draw=black!75,fill=white!75,inner sep=1.5pt,font=\small]
\tikzstyle{pre}=[<-,shorten <=1pt,>=stealth',semithick]
\tikzstyle{post}=[->,shorten <=1pt,>=stealth',semithick]
\node (bush) at (-2.5,0) [entity] {};
\node (laura) at (0, 0) [entity] {}
edge [pre]  node [below,text height=6pt,font=\small] {dbo:spouse}  (bush);
\node (barbara) at (2.5,0) [entity] {}
edge [pre]  node [below,text height=6pt,font=\small] {dbo:children}  (laura)
edge [pre, bend left]  node [below,text height=6pt,font=\small] {dbo:children}  (bush);

\node [above,font=\small] at (bush.north) {$x_1$};
\node [above,font=\small] at (laura.north) {$x_2$};
\node [above,font=\small] at (barbara.north) {$x_3$};
\end{tikzpicture}
\end{minipage}
}

\caption{Examples of predicate path and cycle in DBpedia}
\end{figure}
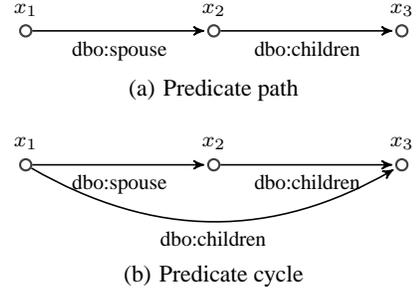

According to the above definitions, predicate paths and predicate cycles can be obtained by replacing entities in paths and cycles with entity variables. Figure~\ref{fig:pred-path} and~\ref{fig:pred-cycle} show the predicate path and predicate cycle corresponding to the path and cycle in Figure~\ref{fig:path} and~\ref{fig:cycle}.

We find that the interesting patterns represented by predicate cycles can be used to infer new facts in KBs. For example, if we have three entities that are connected by any two edges in the predicate cycle shown in Figure~\ref{fig:pred-cycle}, a new fact identified by the third edge can be inferred. A more straightforward way is to generate inference rules from predicate cycles. As an example, the following two rules can be generated from the predicate cycle shown in Figure~\ref{fig:pred-cycle}:

\[dbo:spouse(x_1,x_2)\land dbo:children(x_2,x_3)\]\[\Rightarrow dbo:children(x_1,x_3)\]
\[dbo:children(x_1,x_3)\land dbo:children(x_2,x_3)\]\[\Rightarrow dbo:spouse(x_1,x_2)\]

Based on the above observation, we propose to learn inference rules by finding predicate cycles. However, not all predicate cycles can generate reliable rules, because some of them may present rare connection pattern among entities. Intuitively, the more frequent a predicate cycle occurs in the RDF graph, the more reliable and useful the pattern of the predicate cycle is. Therefore, we define the concept of \textit{Frequent Predicate Path/Cycle}.

\begin{definition}[Frequent Predicate Path/Cycle]
  \label{def:fpp}
A\\ path (cycle) is called the instance of a predicate path (cycle) if it can be generated by instantiating variables with entities in the predicate path (cycle). For a predicate path (cycle), the number of its instances that exists in the given RDF KB is called the support of it. If the support of a predicate path (cycle) is not less than a specified threshold, it is called the frequent predicate path (cycle).
\end{definition}

Frequent predicate cycles (FPCs) are patterns that frequently appear in the KB, rules generated from FPCs are prone to be reliable. So our proposed approach RDF2Rules first mines frequent predicate cycles from RDF KBs, and then generates inference rules from FPCs.

\section{Frequent Predicate Cycle Mining}
This section presents our proposed FPC mining algorithm. The basic idea is first to find all the Frequent Predicate Paths (FPPs) of specified maximum length, and then to discover FPCs by checking which FPPs can form predicate cycles. One big challenge here is that there are huge number of FPP candidates even in a small-sized RDF KB. If there are $N$ predicates, then the number of all possible $k$-predicate paths  (i.e. paths having $k$ predicates) is $(2N)^k$. As an example, there will be 8 million 3-predicate paths if we have 100 predicates. Because counting the supports of predicate paths is the most time-consuming work in the mining process, we have to prune the searching space of predicate paths if we want to find all the FPCs in a reasonable time.

Here we use similar searching strategy in association rule mining and frequent subgraph mining algorithms. The basic idea is to first find frequent $1$-predicate paths, and then iteratively find frequent $k$-predicate paths from frequent ($k$-1)-predicate paths.
Algorithm~\ref{alg:fpm} outlines the proposed algorithm. Our algorithm enumerates every different predicates in the KB, and searches predicate paths that start from it. For a starting predicate, two $1$-predicate paths are first generated and evaluated to find whether they are frequent. After that, our algorithm starts a loop (line 12-20 in Algorithm~\ref{alg:fpm}) that discovers frequent $k$-predicate paths by extending frequent ($k$-1)-predicate paths iteratively. In each iteration, once the frequent $k$-predicate paths are found, FPCs are discovered from them.

\begin{algorithm}[!htb]
\KwIn{An RDF graph $G=(V, P, T)$, the minimum support count $\tau$, maximum length $\xi$}
\KwOut{Frequent predicate cycles $\Theta$\\ \ }
Set $\Theta=\emptyset$;\\
Execute in parallel:\\
\For{each predicate $p_i\in P$}{
Let $\Psi_1=\emptyset$\\
Generate 1-predicate paths $\theta_1=(p_i^{+})$, $\theta_2=(p_i^{-})$;\\
\If{$sup(\theta_1)\ge \tau$}{
$\Psi_1=\Psi_1 \cup \{\theta_1\}$;\\
}
\If{$sup(\theta_2)\ge \tau$}{
$\Psi_1=\Psi_1 \cup \{\theta_2\}$;\\
}
\For{$j=2,...,\xi$}{
Let $\Psi_{j}=\emptyset$;\\
\For{each path $\theta\in \Psi_{j-1}$}{
$\Psi_{j}^{\prime} = pathGrowth(\theta,\tau)$;\\
$\Psi_{j}=\Psi_{j}\cup \Psi_{j}^{\prime}$;\\
}
$\Theta_j= findCycles(\Psi_j)$;\\
$\Theta=\Theta\cup \Theta_j$
}
}
\Return $\Theta$
\caption{Frequent predicate cycles mining algorithm}
\label{alg:fpm}
\end{algorithm}

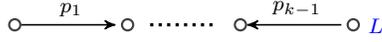
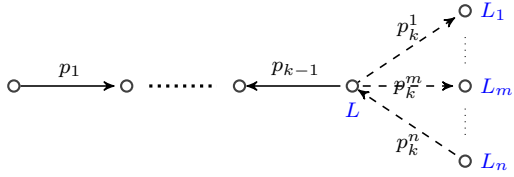
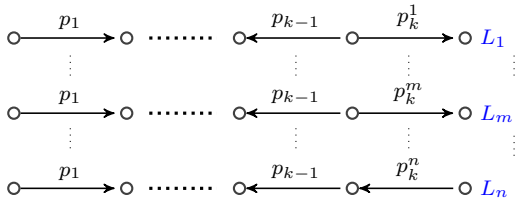
\begin{figure}[!h]
\label{fig:patternGrowth}
\centering
\subfigure[Original ($k$-1)-predicate path]{
\begin{minipage}[b]{0.5\textwidth}
\label{fig:pg-a}
\centering
\begin{tikzpicture}
\tikzstyle{entity}=[circle,thick,draw=black!75,fill=white!75,inner sep=1.5pt,font=\small]
\tikzstyle{pre}=[<-,shorten <=2pt,>=stealth',semithick]
\tikzstyle{post}=[->,shorten <=2pt,>=stealth',semithick]
\node (e1) at (0,0) [entity] {};
\node (e2) at (1.5,0) [entity] {}
edge [pre]  node [above,text height=6pt,font=\small] {$p_{1}$}  (e1);
\node (e3) at (3,0) [entity] {};
\draw[dotted,very thick] (1.8,0) -- (2.7,0);
\node (e4) at (4.5,0) [entity] {}
edge [post]  node [above,text height=6pt,font=\small] {$p_{k-1}$}  (e3);
\node [blue,right,text height=6pt,font=\small] at (e4.east) {$L$};

\end{tikzpicture}

\end{minipage}
}
\subfigure[Extending predicate path]{
\begin{minipage}[b]{0.5\textwidth}
\label{fig:pg-b}
\centering
\begin{tikzpicture}
\tikzstyle{entity}=[circle,thick,draw=black!75,fill=white!75,inner sep=1.5pt,font=\small]
\tikzstyle{pre}=[<-,shorten <=2pt,>=stealth',semithick]
\tikzstyle{post}=[->,shorten <=2pt,>=stealth',semithick]
\node (e1) at (0,0) [entity] {};
\node (e2) at (1.5,0) [entity] {}
edge [pre]  node [above,text height=6pt,font=\small] {$p_{1}$}  (e1);
\node (e3) at (3,0) [entity] {};
\draw[dotted,very thick] (1.8,0) -- (2.7,0);
\node (e4) at (4.5,0) [entity] {}
edge [post]  node [above,text height=6pt,font=\small] {$p_{k-1}$}  (e3);
\node [blue,below,text height=6pt,font=\small] at (e4.south) {$L$};
\node (e5) at (6,1) [entity] {}
edge [pre,dashed]  node [above,text height=6pt,font=\small] {$p_{k}^1$}  (e4);
\node (e6) at (6,0) [entity] {}
edge [pre,dashed]  node [text height=6pt,font=\small] {$p_{k}^m$}  (e4);
\node (e7) at (6,-1) [entity] {}
edge [post,dashed]  node [below,text height=6pt,font=\small] {$p_{k}^n$}  (e4);
\node [blue,right,text height=6pt,font=\small] at (e5.east) {$L_1$};
\node [blue,right,text height=6pt,font=\small] at (e6.east) {$L_m$};
\node [blue,right,text height=6pt,font=\small] at (e7.east) {$L_n$};

\draw[dotted] (6,0.3) -- (6,0.7);
\draw[dotted] (6,-0.3) -- (6,-0.7);
\end{tikzpicture}
\end{minipage}
}

\subfigure[New $k$-predicate paths]{
\begin{minipage}[b]{0.5\textwidth}
\label{fig:pg-c}
\centering
\begin{tikzpicture}
\tikzstyle{entity}=[circle,thick,draw=black!75,fill=white!75,inner sep=1.5pt,font=\small]
\tikzstyle{pre}=[<-,shorten <=2pt,>=stealth',semithick]
\tikzstyle{post}=[->,shorten <=2pt,>=stealth',semithick]
\node (e1) at (0,1) [entity] {};
\node (e2) at (1.5,1) [entity] {}
edge [pre]  node [above,text height=6pt,font=\small] {$p_{1}$}  (e1);
\node (e3) at (3,1) [entity] {};
\draw[dotted,very thick] (1.8,1) -- (2.7,1);
\node (e4) at (4.5,1) [entity] {}
edge [post]  node [above,text height=6pt,font=\small] {$p_{k-1}$}  (e3);
\node (e5) at (6,1) [entity] {}
edge [pre]  node [above,text height=6pt,font=\small] {$p_{k}^1$}  (e4);
\node [blue,right,text height=6pt,font=\small] at (e5.east) {$L_1$};
\node (a1) at (0,0) [entity] {};
\node (a2) at (1.5,0) [entity] {}
edge [pre]  node [above,text height=6pt,font=\small] {$p_{1}$}  (a1);
\node (a3) at (3,0) [entity] {};
\draw[dotted,very thick] (1.8,0) -- (2.7,0);
\node (a4) at (4.5,0) [entity] {}
edge [post]  node [above,text height=6pt,font=\small] {$p_{k-1}$}  (a3);
\node (a5) at (6,0) [entity] {}
edge [pre]  node [above,text height=6pt,font=\small] {$p_{k}^m$}  (a4);
\node [blue,right,text height=6pt,font=\small] at (a5.east) {$L_m$};
\node (b1) at (0,-1) [entity] {};
\node (b2) at (1.5,-1) [entity] {}
edge [pre]  node [above,text height=6pt,font=\small] {$p_{1}$}  (b1);
\node (b3) at (3,-1) [entity] {};
\draw[dotted,very thick] (1.8,-1) -- (2.7,-1);
\node (b4) at (4.5,-1) [entity] {}
edge [post]  node [above,text height=6pt,font=\small] {$p_{k-1}$}  (b3);
\node (b5) at (6,-1) [entity] {}
edge [post]  node [above,text height=6pt,font=\small] {$p_{k}^n$}  (b4);

\node [blue,right,text height=6pt,font=\small] at (b5.east) {$L_n$};
\draw[dotted] (0.75,0.5) -- (0.75,0.8);
\draw[dotted] (0.75,-0.2) -- (0.75,-0.5);
\draw[dotted] (3.75,0.5) -- (3.75,0.8);
\draw[dotted] (3.75,-0.2) -- (3.75,-0.5);
\draw[dotted] (5.25,0.5) -- (5.25,0.8);
\draw[dotted] (5.25,-0.2) -- (5.25,-0.5);
\draw[dotted] (6.65,0.5) -- (6.65,0.8);
\draw[dotted] (6.65,-0.3) -- (6.65,-0.6);
\end{tikzpicture}
\end{minipage}
}

\caption{Illustration of predicate path growth process}
\end{figure}

\begin{algorithm}[t]

\KwIn{An RDF graph $G=(V, P, T)$, a ($k$-1)-predicate path $\theta$, the minimum support count $\tau$}
\KwOut{A set of frequent $k$-predicate path $\Psi$\\ \ }
$\Psi=\emptyset$;\\
\For{each entity $e\in L(\theta)$}{
\For{each edge $(e,p,e^{'})\in T$}{
Generate a new $k$-predicate path $\theta^{'}=(\theta,p^{1})$;\\
\eIf{$\theta^{'}\in \Psi$}{
$\theta^{'}_{\cdot}count=\theta^{'}_{\cdot}count+1$;\\
}{
$\theta^{'}_{\cdot}count=1$;\\
$\Psi=\Psi \cup \{\theta^{'}\}$;\\
}
}
\For{each edge $(e^{'},p,e)\in E$}{
Generate a new $k$-predicate path $\theta^{'}=(\theta,p^{-1})$;\\
\eIf{$\theta^{'}\in \Psi$}{
$\theta^{'}_{\cdot}count=\theta^{'}_{\cdot}count+1$;\\
}{
$\theta^{'}_{\cdot}count=1$;\\
$\Psi=\Psi \cup \{\theta^{'}\}$;\\
}
}
}
\For{each predicate path $\theta^{'}\in \Psi$}{
\If{$\theta^{'}_{\cdot}count<\tau$}{
$\Psi=\Psi/\{\theta^{'}\}$;\\
}
}
\Return $\Psi$
\caption{Predicate path growth algorithm ($pathGrowth$)}
\label{alg:fpg}
\end{algorithm}

The function $pathGrowth$ in Algorithm~\ref{alg:fpm} (line 15) extends a frequent ($k$-1)-predicate path to a set of frequent $k$-predicate paths by adding new predicates to the end of the ($k$-1)-predicate path. Figure 4 illustrates the process of $pathGrowth$. Given a ($k$-1)-predicate path $\theta$ in Figure~\ref{fig:pg-a}, the instance paths of it are first find in the KB, let $L(\theta)$ denotes the set of last entities in the instance paths of $\theta$. Predicates that connect entities in $L(\theta)$ to other entities are then added to the original ($k$-1)-predicate path, as shown in Figure~\ref{fig:pg-b}. In order to prune the search space and always get useful predicate paths, each time a new predicate is added to the original predicate path $\theta$, the following conditions should be satisfied:
\begin{itemize}
\item The added predicate should appear in the frequent $1$-predicate paths (i.e. the added predicate itself is frequent).
\item If the last predicate of the ($k$-1)-predicate path is $p_i^1$, the added predicate can not be $p_i^{-1}$; if the last predicate of the ($k$-1)-predicate path is $p_i^{-1}$, the added predicate can not be $p_i^{1}$. This constraint can eliminate meaningless predicate paths like $(x_1,p_i^{1},x_2,p_i^{-1},x_3,p_i^{1},....)$.
\item The added predicate should connect a required number (i.e. the minimum support) of entities to entities in $L(\theta)$;
\end{itemize}

After adding new predicates, a number of new $k$-predicate paths are obtained, as shown in Figure~\ref{fig:pg-c}. Only the new predicate paths that are also frequent are kept as the output of the function $pathGrowth$.
In the process of extending predicate paths, the set of last entities in the instance paths are always kept, which facilitates adding new predicates and counting the support of predicate paths. Algorithm~\ref{alg:fpg} outlines the detailed steps of the $pathGrowth$ algorithm.

The function $findCycles$ in Algorithm~\ref{alg:fpm} (line 18) finds a set of FPCs from the discovered FPPs. For a FPP $\theta=(x_1,p_{i1}^{d_1},...,x_{k+1})$, the following steps are performed to decide whether a FPC can be generated from it:
\begin{itemize}
\item Find the set of instance paths $I_{\theta}$ of $\theta$ in the KB, set the support of $\theta$ as a cycle $sup_{cycle}(\theta)=0$.
\item For each path $\phi$ in $I_{\theta}$, if the first entity and the last entity in path $\phi$ are the same, $sup_{cycle}(\theta)=sup_{cycle}(\theta)+1$.
\item If $sup_{cycle}(\theta)$ is not less than the minimum support threshold, then a predicate cycle $\theta^{\prime}=(x_1,p_{i1}^{d_1},...,p_{ik}^{d_k},x_{1})$ is generated by changing the last entity variable $x_{k+1}$ to $x_{1}$ in $\theta$; $\theta^{\prime}$ is kept as a FPC.
\end{itemize}

After performing the above steps for each discovered FPP, the $findCycles$ function obtains a set of FPCs, which is returned as the result of FPC mining algorithm.

In order to accelerate the mining process, our algorithm can run parallelly in a multi-core machine. The steps from line 3 to line 21 in Algorithm~\ref{alg:fpm} can be executed independently for each starting predicate. If our algorithm runs on a machine with $m$ cores, $m$ threads can be created to find paths starting with $m$ different predicates in parallel.
\\

\noindent\textbf{Discussion of the Support Measure}

According to Definition~\ref{def:fpp}, the support of a predicate path equals to the number of its instance paths. This standard support measure, however, does not meet the Downward-Closure Property in frequent pattern mining, which requires that the support of a pattern must not exceed that of its sub-patterns. But in the problem of FPC mining, the support of a $k$-predicate path can be larger than that of its sub predicate paths. It is because many predicates in an RDF KB are not functional or inverse-functional\footnote{For functional or inverse-functional predicates, please refer to http://www.w3.org/TR/owl-ref/\#FunctionalProperty-def}. For example, $dbo:children$ predicate in DBpeida is not functional, one people can have more than one child; so it is very likely that predicate path $(x_1,dbo:spouse^1,x_2,dbo:children^1,x_3)$ has more instance paths than $(x_1,dbo:spouse^1,x_2)$ does.

Therefore, extending FPPs by adding new frequent predicates can not ensure safely pruning of infrequent predicate paths. Some FPPs will also be pruned if we determine a predicate path is frequent or not based on the support measure in Definition~\ref{def:fpp}. Actually, our approach can ensure finding all the FPPs if \textit{frequent} is defined based on the following support measure:
\begin{equation}
sup_{var}(\theta)=min_{i\in\{1,...,k+1\}}(|\Pi_{x_i}(\theta)|)
\end{equation}
where $\theta=(x_1,p_1^{d_1},x_2,...,p_k^{d_k},x_{k+1})$ is a k-predicate path, and $\Pi_{x_i}(\theta)$ be the set of entities that instantiate variable $x_i$ in the instance paths of $\theta$. It is easy to find that $sup_{var}$ satisfies the Downward-Closure Property; let $sup(\theta)$ be the standard support defined in Definition~\ref{def:fpp}, then we have $sup_{var}(\theta)\le sup(\theta)$.

By the above discussion, we show that our approach actually finds FPPs having $sup_{var}$ no less than a given threshold; the discovered FPPs are also frequent in terms of the standard support measure defined in Definition~\ref{def:fpp}.

\noindent\textbf{Dealing with Duplicate FPCs}

In the mining process of our approach, the same FPC may be generated from different FPPs. For example, given the following four different predicate paths:
\[(x_1,p_1^{1},x_2,p_2^{1},x_3), (x_1,p_2^{1},x_2,p_1^{1},x_3)\]
\[(x_1,p_1^{-1},x_2,p_2^{-1},x_3), (x_1,p_2^{-1},x_2,p_1^{-1},x_3)\]
we can generate the following predicate cycles from them:
\[(x_1,p_1^{1},x_2,p_2^{1},x_1), (x_1,p_2^{1},x_2,p_1^{1},x_1)\]
\[(x_1,p_1^{-1},x_2,p_2^{-1},x_1), (x_1,p_2^{-1},x_2,p_1^{-1},x_1)\]
Actually, they represent the same predicate cycle shown in Figure 5. To avoid produce duplicate rules, duplicate FPCs should be first detected and eliminated. So a predicate cycle normalization method is used to ensure one predicate cycle can have only one unique representation. To achieve that, we first assign each predicate in the KB a unique id. Then for a predicate path, the predicate $p_{min}$ having the minimum id in it is found and taken as the first predicate in the representation, and the direction of $p_{min}$ is set to $1$; the second predicate is the next one follows the forward direction of $p_{min}$, and so forth. Assuming the subscript index of a predicate is its id, the predicate cycle in Figure 5 has a unique representation as $(x_1,p_1^{1},x_2,p_2^{1},x_1)$ according to our normalization method.

It is very important to remove duplicate FPCs before generating rules. Because evaluating rules is a time consuming task, excluding redundant FPCs enables our approach to run more efficiently.

\begin{figure}
  \centering
  \begin{tikzpicture}
  \tikzstyle{entity}=[circle,thick,draw=black!75,fill=white!75,inner sep=1.5pt,font=\small]
  \tikzstyle{pre}=[<-,shorten <=1pt,>=stealth',semithick]
  \tikzstyle{post}=[->,shorten <=1pt,>=stealth',semithick]
  \node (x1) at (0,0) [entity] {};
  \node (x2) at (4, 0) [entity] {}
  edge [pre, bend right]  node [below,text height=6pt,font=\small] {$p_1$}  (x1)
  edge [post, bend left]  node [below,text height=6pt,font=\small] {$p_2$}  (x1);
  \end{tikzpicture}
  \label{fig:dupcycle}
  \caption{A 2-predicate cycle}
\end{figure}
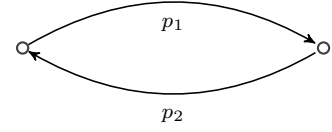

\section{Rule Generation and Evaluation}
\subsection{Generate Rules from FPCs}
Once all the FPCs $\Theta=\{\theta_1,...,\theta_n\}$ are found by Algorithm~\ref{alg:fpm}, rules are generated from them. As discussed in Section~\ref{sec:fpc}, one FPC having $k$ predicates can generates $k$ rules. Formally, for a FPC $\theta=(x_1,p_{1}^{d_1},x_2,p_{2}^{d_2},...,p_{k}^{d_k},x_{1})$, the $j$th rule generated from it is
\begin{equation}
  \label{eq:genrule}
\bigwedge_{i\in [1,k],i\ne j} \langle x_i,p_{i}^{d_i},x_{i+1} \rangle \Rightarrow \langle x_j,p_{j}^{d_j},x_{j+1} \rangle
\end{equation}

For example, a FPC $(x_1,hasParent^{1},x_2,hasChildren^{-1},x_1)$ can generate the following two rules:

\[\langle x_1,hasParent,x_2 \rangle \Rightarrow \langle x_2,hasChildren,x_1 \rangle\]
\[\langle x_2,hasChildren,x_1 \rangle \Rightarrow \langle x_1,hasParent,x_2 \rangle\]

In some cases, one predicate may appear multiple times in a FPC, duplicate rules might be generated from it. In the example shown in Figure~\ref{fig:pred-cycle}, we can generate three rules from this $3$-predicate cycle, but two of them are logically the same. Our approach will detect this problem and filter out duplicate rules.

\subsection{Add Type Information}
Adding entity type information to rules can produce more accurate rules. For example, without entity type information, we may get rules like $\langle x_1, bornIn, x_2 \rangle \Rightarrow \langle x_1,diedIn,x_2 \rangle$. Based on this rule, if we already know that someone was born in someplace, then we can predict that this man also died in the same place. However, if the entity instantiating $x_2$ is a small town, the prediction of this rule is prone to be incorrect, because most people would not stay in the same town in their whole lives. If the entity instantiating $x_2$ becomes a country, the prediction will have a higher probability to be correct. Based on this observation, the following rule is more preferable.
\[ \langle x_1, bornIn, x_2 \rangle \land \langle x_1, typeOf, People \rangle\]\[ \land  \langle x_2, typeOf, Country \rangle \Rightarrow \langle x_1,dieIn,x_2 \rangle \]

To generate rules with entity type information, we propose an algorithm to find frequent types for FPCs and adds type information in FPCs, which is outlined in Algorithm~\ref{alg:addtype}. Given a FPC $\theta$, our algorithm first finds the frequent types of entities for each variable $x_i$ (line 4-17 in Algorithm~\ref{alg:addtype}). Then triples like $\langle x_i, typeOf, type\rangle$ are generated as type constraint for variable $x_i$ (line 21-25 in Algorithm~\ref{alg:addtype}). At last, type constraints for all the variables are combined and added to the original FPC $\theta$, resulting in a number of new FPCs having type constraints (line 27-30 in Algorithm~\ref{alg:addtype}). Two points in our algorithm should be noticed:
\begin{itemize}
\item A $\langle x_i, typeOf, Thing\rangle$ triple is taken as a possible type constraint for $x_i$ in default; $Thing$ is the most general type, any entity is an instance of $Thing$. If this constraint is selected and put over $x_i$, it actually makes no different for $x_i$.
\item For each variable $x_i$, there could be lots of frequent types, especially when many entities have more than one type. In order to avoid generating too many constraint combinations, we only keep the top-$k$ frequent types for each variable. $k$ is a parameter and needs to be set manually.
\end{itemize}

\begin{algorithm}[t]

\KwIn{A FPC $\theta=(x_1,p_{1}^{d_1},x_2,p_{2}^{d_2},...,p_{k}^{d_k},x_{1})$, the minimum support count $\tau$, the maximum number of kept types for each variable $k$}
\KwOut{A set of FPCs with type information $\bar{\Theta}$\\ \ }
Set $\bar{\Theta}=\emptyset$;\\
Get all the instance cycles of $\theta$;\\
\For{$i=1,...,k$}{
Set $C_i=\emptyset$;\\
Get $\Pi_{x_i}(\theta)$, the set of entities instantiating variable $x_i$\\
\For{each entity $e\in \Pi_{x_i}(\theta)$}{
Get the set of types $Type(e)$ of $e$\\
\For{each type $t\in Type(e)$}{
\eIf{$t\in C_i$}{
$t.count=t.count+1$;
}{
$t.count=1$;\\
$C_i=C_i\cup\{t\}$;
}
}
}
Remove types from $C_i$ whose counts are less than $\tau$;\\
\If{$|C_i|>k$}{
Only keep the top-$k$ frequent types in $C_i$;
}
Set $H_i=\{\langle x_i, typeOf, Thing\rangle\}$;\\
\For{each type $t\in C_i$}{
Genenrate a triple $h=\langle x_i, typeOf, t\rangle$;\\
$H_i=H_i\cup\{h\}$;
}
}

\For{each $\bar{h}\in H_1\times H_2 \times ...\times H_k $}{
Generate a FPC with type information $\bar{\theta}=\theta\oplus \bar{h}$;\\
$\bar{\Theta}=\bar{\Theta}\cup \{\bar{\theta}\}$;
}

\Return $\bar{\Theta}$
\caption{Add type information to FPCs}
\label{alg:addtype}
\end{algorithm}

By using Algorithm~\ref{alg:addtype}, we may get a FPC with type information as shown in Figure 6. When we generate rules from a FPC with type information, Equation~\ref{eq:genrule} is first used to get rules without type information; then type constraints in the FPC are added in the body of each rule. For example, the FPC in Figure 6 can produce the rule which is previously presented in this section. In RDF2Rules, rules with and without type information will all be generated.

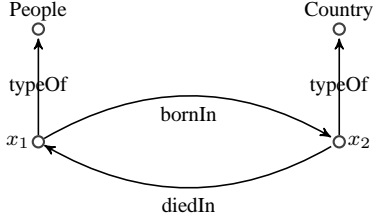
\begin{figure}
  \centering
  \begin{tikzpicture}
  \tikzstyle{entity}=[circle,thick,draw=black!75,fill=white!75,inner sep=1.5pt,font=\small]
  \tikzstyle{pre}=[<-,shorten <=1pt,>=stealth',semithick]
  \tikzstyle{post}=[->,shorten <=1pt,>=stealth',semithick]
  \node (x1) at (0,0) [entity] {};
  \node (x2) at (4, 0) [entity] {}
  edge [pre, bend right]  node [below,text height=6pt,font=\small] {bornIn}  (x1)
  edge [post, bend left]  node [below,text height=6pt,font=\small] {diedIn}  (x1);
  \node (type1) at (0,1.5) [entity] {}
  edge [pre] node [text height=6pt,font=\small] {typeOf} (x1);
  \node (type2) at (4,1.5) [entity] {}
  edge [pre] node [text height=6pt,font=\small] {typeOf} (x2);

  \node [left,font=\small] at (x1) {$x_1$};
  \node [right,font=\small] at (x2) {$x_2$};
  \node [above,font=\small] at (type1) {People};
  \node [above,font=\small] at (type2) {Country};

  \end{tikzpicture}
  \label{fig:fpcschema}
  \caption{Predicate cycle with schema information}
\end{figure}

\subsection{Rule Evaluation}
The measure of support defined in Section 2 measures how frequent a pattern appears in the concerned KB. The support of a rule is equal to the support of its corresponding FPC, i.e. the number of all possible instantiations of the variables in the rule. However, rules with high support may also make inaccurate predications. To evaluate how accurate a rule is, we can use the confidence measure which is used in association rule mining.

\begin{equation}
conf(R_{body} \Rightarrow R_{head})=\frac{sup(R_{body} \Rightarrow R_{head})}{sup(R_{body})}
\end{equation}

If we use this standard confidence to evaluate rules in RDF2Rules, it will treat all that facts (entity relations) that do not exist in the given KB as false ones. It is not suitable for the scenario of enriching knowledge in KBs, because KBs are incomplete and unknown facts can not be simply taken as incorrect. In order to evaluate rules under the \textit{Open World Assumption (OWA)}, AMIE~\cite{galarraga2013amie} uses a PCA confidence measure, which is defined as Equation~\ref{equ:pca}.

\begin{equation}
\label{equ:pca}
\begin{split}
conf_{pca}&(R_{body} \Rightarrow \langle x,p,y \rangle)\\
&=\frac{sup(R_{body} \Rightarrow \langle x,p,y \rangle)}{sup_{pca}(R_{body} \land \langle x,p,y \rangle)}
\end{split}
\end{equation}
where
\begin{equation}
\label{equ:pcasup}
\begin{split}
sup_{pca}&(R_{body} \land \langle x,p,y \rangle)\\
&=|\{ (x,y)|\exists z_1,...,z_m,y^{'}:(R_{body})\land \langle x,p,y \rangle \}|.
\end{split}
\end{equation}
PCA confidence is based on the assumption that if an entity $x$ has relation $p$ with other entities, i.e. $\langle x, p, y \rangle \in KB$, $y\in Y$, then all the relations $p$ of entity $x$ are contained in the KB. If a relation $\langle x, p, y^{'} \rangle$ is predicted while $y^{'}\notin Y$, the new predicted relation will be treated as a false fact. If there is no relation $p$ of $x$ in the KB, then any predicated relation $p$ of $x$ is considered as a true fact. This assumption is also adopted in KnowledgeVault~\cite{dong2014knowledge}. The PCA confidence works well for rules having function predicate in the rule heads, and also holds for predicates having high functionality~\cite{galarraga2013amie}. However, PCA confidence does not work well in some cases. For example, the following rule is mined by AMIE on YAGO2\footnote{This rule is from the webpage of AMIE, http://resources.mpi-inf.mpg.de/yago-naga/amie/}.

\[\langle x_1, livesIn, x_2 \rangle \land \langle x_2, isLocatedIn, x_3 \rangle\]
\[\Rightarrow \langle x_1, isPoliticianOf, x_3 \rangle\]
According to the evaluation results on the webpage of AMIE, this rule has a 85.38\% PCA confidence, but only 13.33\% predictions of this rule is correct. This is because that only a small number of people are politicians, it is not accurate to consider all the predicted $isPoliticianOf$ relations of an entity $e$ as true when $e$ has no $isPoliticianOf$ relations in the origin KB. Actually, if an entity is an instance of class $Politician$, then a new predicted $isPoliticianOf$ relation about this entity is more likely to be true; otherwise, the new predicted fact might be wrong. Therefore, if we can estimate the probability of an entity having a specific relation of predicate, the confidence can be more accurately evaluated.

Let $P(e,p)$ denote the probability of entity $e$ having a relation specified by $p$, we use the entity type information to estimate it, which is computed as
\begin{equation}
\label{eq:prob}
P(e,p) =max_{c\in C_e}\frac{|Inst_p(c)|}{|Inst(c)|}
\end{equation}
where $C_e$ is the set of types of $e$ (one entity can have more than one types), $Inst(c)$ is the set of instances of $c$, and $Inst_p(c)$ is the set of instances of $c$ that have relations of $p$. It is easy to find that $0\le P(e,p)\le 1$. Based on Equation~\ref{eq:prob}, we define a new confidence measure called \textit{soft confidence}

\begin{equation}
  \begin{split}
    conf_{st}&(R_{body} \Rightarrow \langle x,p,y \rangle)\\
    &=\frac{sup(R_{body} \Rightarrow \langle x,p,y \rangle)}{sup(R_{body})-\sum_{e\in U}P(e,p)}
  \end{split}
\end{equation}
where $U$ is the set of entities that previously have no relations of $p$, but have new predicted relations of $p$ by the rule. This new confidence can be computed when the entity type information is available, which can evaluate rules more accurately. We will compare our new confidence with both standard confidence and PCA confidence in the experiments.

\section{RDF Indexing for Mining Algorithm}
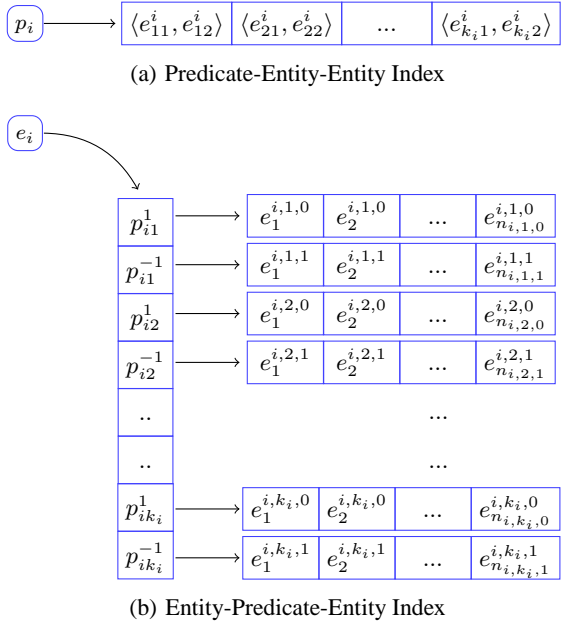
\begin{figure}
\label{fig:peeindex}
\centering
\subfigure[Predicate-Entity-Entity Index]{
\begin{minipage}[b]{0.5\textwidth}
\label{fig:index-a}
\centering
\begin{tikzpicture}
\tikzstyle{entity}=[rectangle,draw=blue!75,fill=white!100,text height=6pt,font=\small,rounded corners]
\tikzstyle{cell}=[rectangle,draw=blue!75]
\tikzstyle{cell2}=[rectangle]
\tikzstyle{space1}=[minimum height=2em,matrix of nodes,row sep=-\pgflinewidth,column sep=-\pgflinewidth,column 1/.style={font=\ttfamily},text depth=0.5ex,text height=2ex,nodes in empty cells]
\tikzstyle{space2}=[minimum height=1.5em,matrix of nodes,row sep=-\pgflinewidth,column sep=-\pgflinewidth,column 1/.style={font=\ttfamily},text depth=0.5ex,text height=2ex,nodes in empty cells]

\node (pi) at ( 0,0) [entity] {$p_i$};

\matrix (pindex1) [space2,row 1/.style={nodes={cell,minimum width=3.8em}}] at (4.2,0)
{
$\langle e_{11}^{i},e_{12}^{i} \rangle$ \pgfmatrixnextcell $\langle e_{21}^{i},e_{22}^{i} \rangle$ \pgfmatrixnextcell $...$ \pgfmatrixnextcell $\langle e_{k_{i}1}^{i},e_{k_{i}2}^{i} \rangle$ \\
};
\draw [->](0.25,0) to (1.2,0);
\end{tikzpicture}
\end{minipage}
}

\subfigure[Entity-Predicate-Entity Index]{
\begin{minipage}[b]{0.5\textwidth}
\label{fig:index-b}
\centering
\begin{tikzpicture}
\tikzstyle{entity}=[rectangle,draw=blue!75,fill=white!100,text height=6pt,font=\small,rounded corners]
\tikzstyle{cell}=[rectangle,draw=blue!75]
\tikzstyle{cell2}=[rectangle]
\tikzstyle{space1}=[minimum height=2em,matrix of nodes,row sep=-\pgflinewidth,column sep=-\pgflinewidth,column 1/.style={font=\ttfamily},text depth=0.5ex,text height=2ex,nodes in empty cells]
\tikzstyle{space2}=[minimum height=1.5em,matrix of nodes,row sep=-\pgflinewidth,column sep=-\pgflinewidth,column 1/.style={font=\ttfamily},text depth=0.5ex,text height=2ex,nodes in empty cells]

\node (ei) at ( 0,1) [entity] {$e_i$};
\matrix (pindex) [space1,column 1/.style={nodes={cell,minimum width=2.3em}}] at (1.6,-2.4)
{
$p_{i1}^{1}$ \\
$p_{i1}^{-1}$ \\
$p_{i2}^{1}$ \\
$p_{i2}^{-1}$ \\
$..$ \\
$..$ \\
$p_{ik_i}^{1}$ \\
$p_{ik_i}^{-1}$ \\
};

\matrix (pindex1) [space2,row 1/.style={nodes={cell,minimum width=3.2em}}] at (5,-0.1)
{
$e_{1}^{i,1,0}$ \pgfmatrixnextcell $e_{2}^{i,1,0}$ \pgfmatrixnextcell $...$ \pgfmatrixnextcell $e_{n_{i,1,0}}^{i,1,0}$ \\
};

\matrix (pindex2) [space2,row 1/.style={nodes={cell,minimum width=3.2em}}] at (5,-0.75)
{
$e_{1}^{i,1,1}$ \pgfmatrixnextcell $e_{2}^{i,1,1}$ \pgfmatrixnextcell $...$ \pgfmatrixnextcell $e_{n_{i,1,1}}^{i,1,1}$ \\
};

\matrix (pindex3) [space2,row 1/.style={nodes={cell,minimum width=3.2em}}] at (5,-1.4)
{
$e_{1}^{i,2,0}$ \pgfmatrixnextcell $e_{2}^{i,2,0}$ \pgfmatrixnextcell $...$ \pgfmatrixnextcell $e_{n_{i,2,0}}^{i,2,0}$ \\
};

\matrix (pindex4) [space2,row 1/.style={nodes={cell,minimum width=3.2em}}] at (5,-2.05)
{
$e_{1}^{i,2,1}$ \pgfmatrixnextcell $e_{2}^{i,2,1}$ \pgfmatrixnextcell $...$ \pgfmatrixnextcell $e_{n_{i,2,1}}^{i,2,1}$ \\
};

\matrix (pindex5) [space2,row 1/.style={nodes={cell,minimum width=3.2em}}] at (5,-4)
{
$e_{1}^{i,k_i,0}$ \pgfmatrixnextcell $e_{2}^{i,k_i,0}$ \pgfmatrixnextcell $...$ \pgfmatrixnextcell $e_{n_{i,k_i,0}}^{i,k_i,0}$ \\
};

\matrix (pindex6) [space2,row 1/.style={nodes={cell,minimum width=3.2em}}] at (5,-4.65)
{
$e_{1}^{i,k_i,1}$ \pgfmatrixnextcell $e_{2}^{i,k_i,1}$ \pgfmatrixnextcell $...$ \pgfmatrixnextcell $e_{n_{i,k_i,1}}^{i,k_i,1}$ \\
};

\matrix (pindex7) [space2,row 1/.style={nodes={cell2,minimum width=3.2em}}] at (5,-2.7)
{
\pgfmatrixnextcell  \pgfmatrixnextcell $...$ \pgfmatrixnextcell \\
};

\matrix (pindex8) [space2,row 1/.style={nodes={cell2,minimum width=3.2em}}] at (5,-3.35)
{
\pgfmatrixnextcell  \pgfmatrixnextcell $...$ \pgfmatrixnextcell \\
};

\draw [->](0.25,1) to [bend left=30] (1.5,0.25);
\draw [->](2,-0.1) to (2.85,-0.1);
\draw [->](2,-0.75) to (2.85,-0.75);
\draw [->](2,-1.4) to (2.85,-1.4);
\draw [->](2,-2.05) to (2.85,-2.05);
\draw [->](2,-4) to (2.85,-4);
\draw [->](2,-4.65) to (2.85,-4.65);

\end{tikzpicture}
\end{minipage}
}
\caption{Indexes of RDF graph}
\label{fig:index}
\end{figure}

In order to ensure the efficiency of our approach, we propose to use a in-memory indexing structure to support the mining algorithm instead of using the existing RDF storage systems. Generally, there are three types of queries over the RDF graph in our mining algorithm:
\begin{itemize}
\item[(1)] Given a predicate, find all the entity pairs that it connects;
\item[(2)] Given an entity, find all its incident edges and its neighbor entities;
\item[(3)] Given a predicate path, find all of its instances path.
\end{itemize}
The query of the first type is used to generate frequent 1-predicate paths in Algorithm~\ref{alg:fpm}. The second one is used in the predicate path growth process. The third one is used for counting the supports of predicate paths and finding predicate cycles. In order to support the above queries, two indexes are used in our approach, the \textit{Predicate-Entity-Entity} index and the \textit{Entity-Predicate-Entity} index.
\begin{itemize}
\item The \textit{Predicate-Entity-Entity} index uses all the predicates as keys, and entity pairs connected by predicates as valules. Figure~\ref{fig:index-a} shows a general example entry of the \textit{Predicate-Entity-Entity} index; a predicate key $p_i$ is associated to a vector of $k_i$ entity pairs; an entity pair $\langle e_{j1}^{i},e_{j2}^{i} \rangle$ in the vector corresponds to an edge $\langle e_{j1}^{i},p_i,e_{j2}^{i} \rangle$ in the RDF graph. The \textit{Predicate-Entity-Entity} index is used for queries of type 1.
\item The \textit{Entity-Predicate-Entity} index maps each entity to a sub-index, where the keys are the predicates and the values are vectors of entities. Figure~\ref{fig:index-b} shows an example entry of the \textit{Entity-Predicate-Entity} index; an entity key $e_i$ is mapped to a vector of $2k_i$ predicate keys $\langle p_{i1}^{1},p_{i1}^{-1},..., p_{ik_i}^{1},p_{ik_i}^{-1}\rangle$; each predicate key $p_{ij}^{1}$ or $p_{ij}^{-1}$ is linked to a vector of entities, which are connected to $e_i$ by $p_{ij}^{1}$ or $p_{ij}^{-1}$. The \textit{Entity-Predicate-Entity} index is used for queries of type 2 and 3.
\end{itemize}

\section{Experiments}

\subsection{Experiment Setup}
\noindent \textbf{Datasets.} We evaluate our approach on YAGO2 and DBpedia 2014 (English version). YAGO2 is an extension of YAGO, which is built automatically from Wikipedia, GeoNames, and WordNet. DBpedia is a large-scale RDF KB, which is built by extracting structured content from the information contained the Wikipedia. Details of the used datasets are outlined in Table~\ref{tab:datasets}. Entity types of YAGO2 are obtained from YAGO3~\cite{yago3} (the latest version of YAGO), as Gal{\'a}rraga et al. did in their work~\cite{amieplus}. For DBpedia 2014, we use the dataset of mapping based properties, entity types are from the DBpedia ontology. The third column in Table~\ref{tab:datasets} are the numbers of facts excluding the \textit{rdf:type} statements. For each used KB, one dataset without entity types and another dataset with entity types are generated, which are used for AMIE+ separately in experiments. RDF2Rules always takes the dataset with types as input, but whether learning rules with types or not can be controlled.

\begin{table}[!htb]
\begin{center}
\caption{Details of used KBs}
\label{tab:datasets}
\begin{tabular}{|l|r|r|r|r|}
\hline
Knowledge Base & \#Entities & \#Preds & \#Facts & \#Types \\
\hline
\hline
YAGO2 & 834 K & 32 & 948 K & 225 K\\
\hline
DBpedia 2014 & 4.1M & 669 & 14.8 M & 685\\
\hline
\end{tabular}
\end{center}
\end{table}

\noindent\textbf{Settings.} According to the experimental results reported in~\cite{amieplus}, AMIE+ outperforms both AMIE and several state-of-the-art ILP approaches. So we just compare our approach to AMIE+ in the experiments. All the experiments are run on a server with two 6-Core CPUs (Intel Xeon 2.4GHz) and 48 GB RAM, the operation system is Ubuntu 14.04. In all the experiments, the parameter $k$ in Algorithm~\ref{alg:addtype} is set to 1 for RDF2Rules; we set all the thresholds on confidence measures to 0 for RDF2Rules to let it output all the learned rules. For AMIE+, except for the \textit{minimum support} and the \textit{max depth} (i.e. maximum number of predicates) of rules, all its parameters are set to its default values (head coverage threshold $minHC=0.01$, confidence threshold $minConf=0$, PCA confidence threshold $minpca=0$, etc.); the number of threads that AMIE+ uses is set to the actual number of cores.

\subsection{Mining Efficiency Analysis}
In this sub-section, YAGO2 are used to evaluate the efficiency of RDF2Rules. The running time and the number of learned rules of RDF2Rules are compared with those of AMIE+. YAGO2 is a relatively small KB, so we set the support threshold to 50 for both RDF2Rules and AMIE+. We also let both systems run with three different maximum rule lengths, i.e. 2, 3, and 4. The experimental results are outlined in Table~\ref{tab:peryago2}.

\noindent\textbf{Runtime.}
When the entity type information is not taken into account, both AMIE+ and RDF2Rules can learn rules very quickly. As shown in Table~\ref{tab:peryago2}, RDF2Rules runs faster than AMIE+ does; as the maximum length of rules increases, the efficiency superiority of RDF2Rules is more obvious. When the rules of length 2 are learned, AMIE+ takes about 1 more second than RDF2Rules; when the maximum length of rules is 4, AMIE+ takes about twice the time that RDF2Rules takes. This is because RDF2Rules first discovers FPPs and then generates rules, a FPP of length $k$ can generate $k$ rules. So when longer rules are to be learned, RDF2Rules has more advantage over AMIE+ in terms of running time.

When the entity type information is considered in rule learning process, longer time is consumed for both approaches. But for AMIE+, it can not finish mining in 2 days when the maximum length of rules is 3 or 4. And we find that none of the rules learned by AMIE+ contain type constraints, which are the just the same rules as learned on the dataset without entity types. There is not a parameter of AMIE+ that controls using types or not. We just follow the method that is used in~\cite{amieplus} to allow AMIE+ to learn rules with types, i.e. augmenting YAGO2 dataset by adding the \textit{rdf:type} statements. But AMIE+ can not return rules with types in our experiments, adding the \textit{rdf:type} statements can only augment the dataset and slow AMIE+ down. RDF2Rules can get rules with type constraints in acceptable time for different maximum rule lengths. When the maximum length of rule is set to 4, RDF2Rules finds more than 8 thousand rules and 6,585 of them are rules with types. Table~\ref{tab:ruleswithtype} lists some examples of these rules.

\noindent\textbf{Number of Rules.}
As for the number of rules, RDF2Rules always gets more rules than AMIE+ does given the same maximum rule length. This is because AMIE+ uses a measure called head coverage to prune the search space, which requires the rules covering a certain ratio of facts of the predicate in rule head. AMIE+ also has a relation size threshold to filter out rules with small sized rule head (i.e. the number of facts of predicate in the rule head is small). RDF2Rules uses different pruning strategy from AMIE+ does, there is no threshold on head coverage or the size of rule head, so more rules are searched and returned by RDF2Rules. And the extra rules found by RDF2Rules are also useful for predicting new facts.
\\

\begin{table}[htb]
\begin{center}
\caption{Results on YAGO2}
\label{tab:peryago2}
\begin{tabular}{|c|l|r|r|r|r|}
\hline
With Types & Approach & MaxLen. &\#Rules & Time \\
\hline
\hline
\multirow{ 6}{*}{No} & \multirow{ 3}{*}{AMIE+}  & 2 & 29 & 10.77s \\

 \cline{3-5}

& & 3 &  75 & 30.56s \\

 \cline{3-5}

& & 4 & 662 & 11m18s \\

 \cline{2-5}

 & \multirow{ 3}{*}{RDF2Rules} & 2 & 39 & 9.42s \\

  \cline{3-5}

 & & 3 & 210 & 23.51s \\

  \cline{3-5}

 & & 4 & 1766 & 5m35s \\
\hline

\multirow{ 6}{*}{Yes} & \multirow{ 3}{*}{AMIE+}  & 2 & 29 & 1m43s \\

 \cline{3-5}

& & 3 &  75 & >2 days \\

 \cline{3-5}

& & 4 & 662 & >2 days \\

 \cline{2-5}

& \multirow{ 3}{*}{RDF2Rules} & 2 & 148 & 37s \\

 \cline{3-5}

& & 3 & 1237 & 3m15s \\

 \cline{3-5}

& & 4 & 8351 & 52m11s \\
\hline
\end{tabular}
\end{center}
\end{table}

\begin{table*}[htb]
\begin{center}
\caption{Example rules having types learned by RDF2Rules from YAGO2 (rule length$\le 3$)}
\label{tab:ruleswithtype}
\begin{tabular}{|l|}
\hline
$\langle x_1, created, x_2\rangle  \land \langle x_2, typeOf, wordnet\_movie\_106613686\rangle  \land \langle x_1, typeOf, wordnet\_person\_100007846\rangle  $\\$ \Rightarrow  \langle x_1, directed, x_2\rangle$ \\ \hline
$\langle x_1, livesIn, x_2\rangle  \land \langle x_1, typeOf, wikicat\_Living\_people\rangle  \land \langle x_2, typeOf, wordnet\_country\_108544813\rangle  $\\$ \Rightarrow  \langle x_1, isCitizenOf, x_2\rangle$ \\ \hline
$\langle x_1, worksAt, x_2\rangle  \land \langle x_2, typeOf, wordnet\_university\_108286569\rangle  \land \langle x_1, typeOf, wordnet\_scientist\_110560637\rangle $\\$  \Rightarrow  \langle x_1, graduatedFrom, x_2\rangle$ \\ \hline
$\langle x_1, isPoliticianOf, x_2\rangle  \land \langle x_1, diedIn, x_3\rangle  \land  \langle x_1, typeOf, wikicat\_American\_people\rangle $\\$ \land  \langle x_3, typeOf, wordnet\_administrative\_district\_108491826\rangle   \Rightarrow  \langle x_3, isLocatedIn, x_2\rangle$ \\ \hline
$\langle x_1, hasOfficialLanguage, x_2\rangle  \land \langle x_1, typeOf, wordnet\_country\_108544813\rangle  \land \langle x_3, hasOfficialLanguage, x_2\rangle $\\$ \land  \langle x_3, typeOf, wordnet\_country\_108544813\rangle   \Rightarrow  \langle x_3, dealsWith, x_1\rangle$ \\ \hline
$\langle x_1, diedIn, x_2\rangle  \land \langle x_2, typeOf, wordnet\_country\_108544813\rangle  $\\$ \Rightarrow  \langle x_1, wasBornI, x_2\rangle$ \\ \hline
\end{tabular}
\end{center}
\end{table*}

\subsection{Confidence Measure Evaluation}
We propose a new confidence measure call \textit{soft confidence} in Section 4.3. Table~\ref{tab:amierules} and Table~\ref{tab:rdf2rulerules} list some rules learned by AMIE+ and RDF2Rules from YAGO2 respectively. Rules in Table~\ref{tab:amierules} are the top-10 rules with highest PCA confidences, and rules in Table~\ref{tab:rdf2rulerules} are the top-10 rules with highest soft confidence. We find that the second and third rules in Table~\ref{tab:amierules} predicate $isPoliticianOf$ facts; as we discussed in Section 4.3, there might be many people having $diedIn$ and $isLocatedIn$ relations, but only a small number of them are politicians. So most predictions of these two rules would be wrong. Using soft confidence, the above two rules are penalized and there are not in the top-10 list.

In order to further evaluate the effectiveness of our new confidence measure, we let RDF2Rules and AMIE+ learn rules from DBpedia KB; we select the top rules returned by two approaches, and use them to predict new facts. The quality of new predicted facts can indirectly reflect the reliability of confidence measures. To perform the experiment, we randomly selected 40\% triples from DBpedia dataset as test data, and then removed the test triples before the DBpedia data is fed to RDF2Rules and AMIE+. The support threshold is set to 500 for both two approaches, and only rules of length 2 are learned. Since AMIE+ fails to learn rules with types in the former experiments, we only let RDF2Rules learn rules with types. The top-500 rules are selected each time to predict new facts.

Table~\ref{tab:perdbpedia} shows the results. When rules with no types are learned, AMIE+ and RDF2Rules take very close time; but RDF2Rules generates more rules than AMIE+. When rules with types are learned, RDF2Rules takes about 25 minutes and gets more than 8 thousand rules. The sixth column of Table~\ref{tab:perdbpedia} lists the number of predicated new facts by the top-500 rules; and the last column shows that number of new facts that are found in the holdout test data. Although rules of AMIE+ generate more predictions, rules of RDF2Rules get more facts found in the test data. It seems that soft confidence can evaluate rules more precisely, which generate more accurate predictions. Comparing rules without types and with types learned by RDF2Rules, we also find that adding type information in rules results in less predictions but more hits in the test data.

\begin{table*}[htb]
\begin{center}
\caption{Results on DBpedia}
\label{tab:perdbpedia}
\begin{tabular}{|l|r|r|r|r|r|r|}
\hline
Approach & With Types &MaxLen. &\#Rules & Time & \#Predictions & \#Hits  \\
\hline
\hline
AMIE+  & No &2 & 521 & 5m37s & 83 K & 3.6 K \\

\hline
RDF2Rules & No &2 & 2,791 & 5m13s & 75 K & 3.9 K\\
\hline
RDF2Rules & Yes &2 & 8,462 & 25m13s & 67 K & 4.1 K\\
\hline\end{tabular}
\end{center}
\end{table*}

\begin{table}[!htb]
\begin{center}
\caption{Top-10 Rules learned by AMIE+ from YAGO2 (according to PCA confidence, rule length$\le 3$)}
\label{tab:amierules}
\begin{tabular}{|l|}
\hline
$\langle x_1, isMarriedTo, x_2\rangle   \Rightarrow \langle x_2, isMarriedTo, x_1\rangle$ \\ \hline
$\langle x_1, diedIn, x_2\rangle \land   \langle x_2, isLocatedIn, x_3\rangle$ \\  $\Rightarrow \langle x_1, isPoliticianOf, x_3\rangle$ \\ \hline
$\langle x_1, isLocatedIn, x_2\rangle \land   \langle x_3, livesIn, x_1\rangle$ \\  $\Rightarrow \langle x_3, isPoliticianOf, x_2\rangle$ \\ \hline
$\langle x_1, hasOfficialLanguage, x_2\rangle \land   \langle x_3, isLocatedIn, x_1\rangle$ \\ $\Rightarrow \langle x_3, hasOfficialLanguage, x_2\rangle$ \\ \hline
$\langle x_1, isMarriedTo, x_2\rangle \land   \langle x_2, livesIn, x_3\rangle$ \\  $\Rightarrow \langle x_1, livesIn, x_3\rangle$ \\ \hline
$\langle x_1, isMarriedTo, x_2\rangle \land   \langle x_1, livesIn, x_3\rangle$  \\ $\Rightarrow \langle x_2, livesIn, x_3\rangle$ \\ \hline
$\langle x_1, hasOfficialLanguage, x_2\rangle \land   \langle x_1, isLocatedIn, x_3\rangle$ \\ $\Rightarrow \langle x_3, hasOfficialLanguage, x_2\rangle$ \\ \hline
$\langle x_1, created, x_2\rangle \land   \langle x_1, produced, x_2\rangle   \Rightarrow \langle x_1, directed, x_2\rangle$ \\ \hline
\end{tabular}
\end{center}
\end{table}

\begin{table}[!htb]
\begin{center}
\caption{Top-10 Rules learned by RDF2Rules from YAGO2 (according to soft confidence, rule length$\le 3$)}
\label{tab:rdf2rulerules}
\begin{tabular}{|l|}
\hline
$\langle x_1, hasChild, x_2\rangle  \langle x_1, isMarriedTo, x_3\rangle$ \\  $\Rightarrow \langle x_3, hasChild, x_2\rangle$ \\ \hline
$\langle x_1, isMarriedTo, x_2\rangle   \Rightarrow \langle x_2, isMarriedTo, x_1\rangle$ \\ \hline
$\langle x_1, isMarriedTo, x_2\rangle  \langle x_2, livesIn, x_3\rangle$ \\  $\Rightarrow \langle x_1, livesIn, x_3\rangle$ \\ \hline
$\langle x_1, dealsWith, x_2\rangle  \langle x_3, dealsWith, x_1\rangle$  \\ $\Rightarrow \langle x_3, dealsWith, x_2\rangle$ \\ \hline
$\langle x_1, directed, x_2\rangle   \Rightarrow \langle x_1, created, x_2\rangle$ \\ \hline
$\langle x_1, isLocatedIn, x_2\rangle  \langle x_3, livesIn, x_1\rangle$ \\  $\Rightarrow \langle x_3, livesIn, x_2\rangle$ \\ \hline
$\langle x_1, hasChild, x_2\rangle  \langle x_3, hasChild, x_2\rangle$  \\ $\Rightarrow \langle x_1, isMarriedTo, x_3\rangle$ \\ \hline
$\langle x_1, hasOfficialLanguage, x_2\rangle  \langle x_1, isLocatedIn, x_3\rangle$ \\ $\Rightarrow \langle x_3, hasOfficialLanguage, x_2\rangle$ \\ \hline
$\langle x_1, dealsWith, x_2\rangle   \Rightarrow \langle x_2, dealsWith, x_1\rangle$ \\ \hline
$\langle x_1, hasCapital, x_2\rangle  \langle x_3, livesIn, x_2\rangle   \Rightarrow \langle x_3, livesIn, x_1\rangle$ ~~~\\ \hline

\end{tabular}
\end{center}
\end{table}

\section{Related Work}
As mentioned above, AMIE~\cite{galarraga2013amie} is the most related work to our approach. AMIE mainly focuses on how to evaluate rules under the \textit{Open World Assumption}, and its searching strategy becomes inefficient when dealing with large-scale KBs and long rules. AMIE learns one rule at a time by gradually adding new atoms to the rule body, while our approach first mines Frequent Predicate Paths and then generates several rules from each Frequent Predicate Path. Most recently, AMIE has been Extended to AMIE+ by a series of improvements to make it more efficient~\cite{amieplus}. Lots of work has been done in the domain of Inductive Logic Programming to learn first-order Horn clauses from KBs. The learned rules can also be used for inferring new facts in KBs. Typical ILP systems, such as FOIL~\cite{Quinlan:1990} and Progol~\cite{progol}, need a set of training examples that contain both positive and negative examples of the target concepts or relations. Although ILP systems such as~\cite{muggleton1997learning} are proposed to learn rules from only positive examples, the main problem with these approaches is the low efficiency when dealing with large-scale KBs.

Besides using rules to infer new facts in KB, there are also some other methods proposed for enriching an existing KB. Nickel et al. proposed a tensor factorization method RESCAL~\cite{RESCAL} for relational learning; RESCAL represents entities as low dimensional vectors and relations by low rank matrices, which are learned using a collective learning process. RESCAL has been applied to YAGO and it can predict unknown facts with high accuracy~\cite{Fact_YAGO}; and there is also an improved work on RESCAL~\cite{pasquale_icdm2014}. Bordes et al. proposed a series of embedding approaches of KB that predict new facts from the existing ones~\cite{bordes2011se,bordes2014sme,bordes2013transe}; these approaches embed entities and relations in a KB into a continuous vector space while preserving the original knowledge; new facts are predicted by manipulating vectors and matrices. Socher et al.~\cite{NIPS2013_5028} proposed to use neural tensor network for reasoning over relationships between entities in KBs. Both tensor factorization approaches and embedding approaches all need to learn large number of parameters, which is very time consuming; and what these approaches learned is difficult for human experts to understand; however, they can be valuable complementation of the rule based approaches for enriching new facts in KB. Because the goal of this work is to efficiently learn inference rules from KB and accurately evaluate the learned rules, we do not compare our approaches with the above mentioned ones in the experiments.

There are also some work that uses similar structures as predicate path to predict new facts in knowledge bases, such as~\cite{Lao:2011,metapathWWW2015}. But these work focus on how to accurately predicate relations based on multiple predicate/relation paths. How to effectively discover useful predicate paths are not discussed in these work. Our work focus on how to learn frequent predicate paths and use them to generate rules; the paths discovered by our approach can be used as input for the above two approaches. Most recently, there have been some work try to combine logic rules and knowledge embedding to predict new facts, such as~\cite{Wang:2015:ijcai,zhiyuan2015}. These work also do not focus on how to learn rules, but on how to use rules to make accurate predictions. So rules learned by our approach can also used in these approaches.

\section{Conclusion}
In this paper, we propose a novel rule learning approach RDF2Rules for RDF KBs. Rules are learned by finding frequent predicate cycles in RDF graphs. A new confidence measure is also proposed for evaluating the reliability of the mined rules. Experiments show that our approach outperforms the compared approach in terms of both the quality of predictions and the running time.

\section{Acknowledgements}
The work is supported by NSFC (No. 61202246), NSFC-ANR(No. 61261130588), and the Fundamental Research Funds for the Central Universities (2013NT56).

\bibliographystyle{abbrv}
\bibliography{rdf2rules}

\end{document}